\documentclass[journal]{IEEEtran}
\usepackage{graphicx}
\usepackage{cite}
\usepackage{picinpar}
\usepackage{amsmath}
\usepackage{url}
\usepackage{flushend}
\usepackage[latin1]{inputenc}
\usepackage{colortbl}
\usepackage{soul}
\usepackage{multirow}
\usepackage{pifont}
\usepackage{color}
\usepackage{alltt}
\usepackage[hidelinks]{hyperref}
\usepackage{enumerate}
\usepackage{siunitx}
\usepackage{url}
\usepackage{pbox}
\usepackage{booktabs}
\usepackage{amsmath,amssymb}
\usepackage{algorithmicx,algorithm}
\usepackage{float}
\usepackage{hyperref}
\usepackage{threeparttable}
\usepackage{makecell}

\hypersetup{
    colorlinks=true,
    citecolor=blue,
    linkcolor=red,
    urlcolor=blue
}

\begin{document}
\title{Rethinking Transparent Object Grasping: Depth Completion with Monocular Depth Estimation and Instance Mask}

\author{
	\vskip 1em

    Yaofeng Cheng, \IEEEmembership{Student Member, IEEE}, Xinkai Gao, Sen Zhang, Fusheng Zha, \\ Chao Zeng \IEEEmembership{Member, IEEE}, Lining Sun and Chenguang Yang, \IEEEmembership{Fellow, IEEE}

	\thanks{
		Yaofeng Cheng, Xinkai Gao, Sen Zhang, Fusheng Zha, and Lining Sun are with the State Key Laboratory of Robotics and System at Harbin Institute of Technology. Harbin 150001, China. Fusheng Zha is also at Lanzhou University of Technology. (e-mail: \{chengyf, gaoxinkai, zhangsen\}@stu.hit.edu.cn; \{zhafusheng, lnsun\}@hit.edu.cn).
		
		Chao Zeng and Chenguang Yang are with the Department of Computer Science. University of Liverpool. L69 3BX Liverpool, U.K. (e-mail: chaozeng, cyang@ieee.org).
		Yaofeng Cheng finish this work during the academic visiting in University of Liverpool sponsored by China Scholarship Council Awards.
	}
}

\maketitle

\vspace{1em}
\noindent\textbf{Note: This work has been submitted to the IEEE for possible publication. Copyright may be transferred without notice, after which this version may no longer be accessible.}
\vspace{1em}

\begin{abstract}
Due to the optical properties, transparent objects often lead depth cameras to generate incomplete or invalid depth data, which in turn reduces the accuracy and reliability of robotic grasping.
Existing approaches typically input the RGB-D image directly into the network to output the complete depth, expecting the model to implicitly infer the reliability of depth values. 
However, while effective in training datasets, such methods often fail to generalize to real-world scenarios, where complex light interactions lead to highly variable distributions of valid and invalid depth data.
To address this, we propose \textbf{ReMake}, a novel depth completion framework guided by an instance mask and monocular depth estimation. 
By explicitly distinguishing transparent regions from non-transparent ones, the mask enables the model to concentrate on learning accurate depth estimation in these areas from RGB-D input during training. 
This targeted supervision reduces reliance on implicit reasoning and improves generalization to real-world scenarios. 
Additionally, monocular depth estimation provides depth context between the transparent object and its surroundings, enhancing depth prediction accuracy.
Extensive experiments show that our method outperforms existing approaches on both benchmark datasets and real-world scenarios, demonstrating superior accuracy and generalization capability. Code and videos are available at \url{https://chengyaofeng.github.io/ReMake.github.io/}.
\end{abstract}

\def\abstractname{Note to Practitioners}
\begin{abstract}
Transparent objects are often misestimated by depth cameras in real-world scenarios due to reflection and refraction, resulting in unreliable robotic grasps. 
To address this, it is essential to accurately recover depth from the input RGB-D image while ensuring that the method generalizes well under complex lighting conditions.
In this work, we address the challenge of perceiving transparent objects by combining instance segmentation and monocular depth estimation. 
Our method explicitly identifies transparent regions using an instance segmentation mask and leverages depth context between transparent and non-transparent regions to guide depth reconstruction. 
This leads to more accurate depth perception of transparent objects and supports robust robotic grasping in real-world scenarios.
Our method significantly improves depth prediction accuracy from single-view RGB-D input and increases grasp success rates by 53\%-63\% across both bird's-eye and horizontal camera views. 
These results demonstrate strong performance and generalization to real-world conditions. 
In future work, we aim to tackle the unique challenges posed by hollow transparent objects, which continue to hinder reliable robotic perception and manipulation.
\end{abstract}


\markboth{IEEE}%
{}

\definecolor{limegreen}{rgb}{0.2, 0.8, 0.2}
\definecolor{forestgreen}{rgb}{0.13, 0.55, 0.13}
\definecolor{greenhtml}{rgb}{0.0, 0.5, 0.0}

\section{Introduction}

\begin{figure}[!ht]\centering
	\includegraphics[width=3.5in]{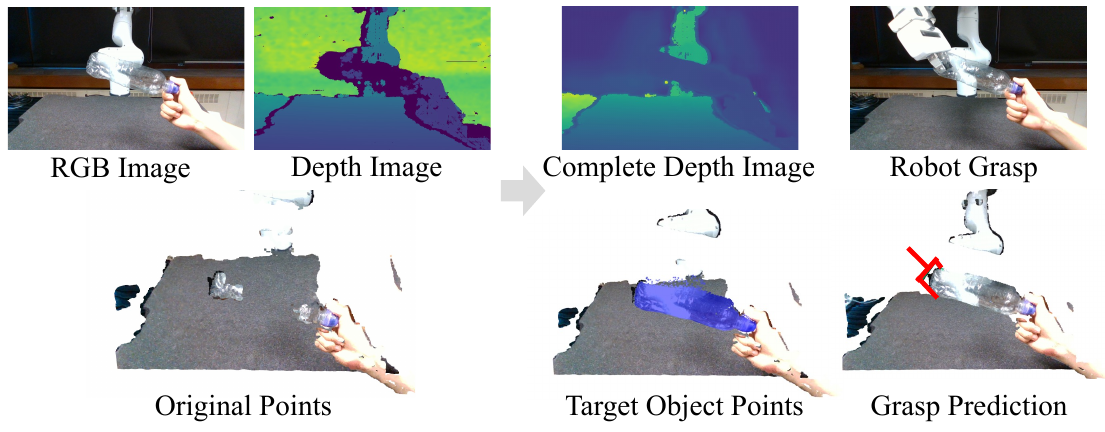}
	\caption{Method Overview. We predict a complete depth map from a single RGB-D image, enabling accurate target object points extraction and 6-DoF grasp prediction.}\label{fig_1}
\end{figure}

\IEEEPARstart{T}{ransparent} objects are ubiquitous in daily life, including laboratories, kitchens, hospitals, and factories. However, widely used depth sensors frequently fail to capture them accurately. Taking RealSense as an example, the infrared light emitted by the camera can be refracted or reflected by the surfaces of transparent objects, leading to incorrect or missing depth values in certain regions. This makes it difficult to reliably grasp such objects. Consequently, accurately predicting the depth of transparent objects remains a key challenge for successful robotic grasping.

Early research on transparent object perception has explored methods to reconstruct depth using multi-view information. These approaches typically capture RGB images \cite{klank2011transparent, dai2023graspnerf, kerr2022evo} or multi-view depth images \cite{zhou2019glassloc, albrecht2013seeing} or both together \cite{ji2017fusing} through camera motion, and then synthesizing information across views to recover the object's actual shape. However, acquiring and fusing multiple images is time-consuming and often impractical in occluded or constrained environments, such as when an object is placed inside a cabinet.

In contrast, recent work on transparent object grasping has focused on more practical single-view methods, where an RGB-D image captured from a fixed viewpoint is input to a network to predict the complete depth map of the target object \cite{sajjan2020clear, fang2022transcg, dai2022domain, tang2021depthgrasp, li2023fdct, zhu2021rgb, zhai2024tcrnet, yan2024transparent}. In these methods, models are trained on datasets that include ground-truth depth annotations, and losses are often computed specifically on transparent regions to improve reconstruction accuracy. However, we observe that despite achieving promising results on these datasets, these methods often fail to generalize well in real-world scenarios, struggling to adapt to new environments and camera viewpoints.

Our analysis of existing datasets and their usage in prior methods (see details in Sec. \ref{analysis_section}), reveals that so-called ``transparent regions'' are not uniform. They include not only missing-depth areas caused by the absence of reflected infrared light, but also refractive regions with inaccurate depth values, as well as normal regions with accurate depth. Although refractive regions contain depth values, these values are often invalid due to distortion (see Fig. \ref{fig_2} (b)). Existing approaches feed an RGB-D image into the network and rely on the model to distinguish between valid and invalid depth values using only image information, in order to produce an accurate target depth map. However, because optical phenomena like refraction and reflection are highly variable, introducing new objects, scenes, or viewpoints often leads to significant changes in the appearance of affected regions (see Fig. \ref{fig_2} (c)). Consequently, relying solely on global image information to distinguish between valid and invalid depth values becomes increasingly difficult, leading to poor generalization in existing methods. This insight motivates us to explore new strategies that reduce the model's learning burden by explicitly guiding depth estimation within transparent regions.

In this paper, we propose a novel framework named \textbf{ReMake} (\textbf{Re}lative depth and \textbf{Ma}s\textbf{k} att\textbf{e}ntion guided depth completion), for completing the depth of transparent objects. Unlike prior methods such as ClearGrasp \cite{sajjan2020clear}, OOD \cite{zhu2021rgb}, and TransAFF \cite{jiang2022a4t}, which utilize masks merely to remove transparent depth values during training, ReMake incorporates the mask directly into the network input to explicitly indicate transparent regions. During training, the model learns to associate masked areas with depth-ambiguous regions, enabling it to focus on predicting depth in these regions using the surrounding RGB-D context. This targeted supervision alleviates the need to infer depth reliability implicitly and enhances generalization to novel objects and scenes.

Moreover, inspired by the human ability to infer object depth through relative positioning of objects \cite{oliva2007role}, we introduce a relative depth map as an additional input channel. We observe that monocular depth estimation methods capture meaningful spatial context \cite{eigen2014depth, fu2018deep, silberman2012indoor, geiger2013vision},  which we incorporate to provide relative depth cues for improved prediction in transparent regions. This provides contextual depth relationships between the transparent object and its surroundings, enabling the network to more effectively estimate depth in complex scenes.

To summarize, our key contributions are as follows:

\begin{enumerate}
	\item We propose a novel transparent object depth prediction framework that uses a mask to distinguish transparent regions from the background. This explicit guidance directs the network's attention toward estimating depth in these challenging areas, leading to improved generalization in real-world scenes. 
	
	\item We introduce the use of relative depth maps to assist in reconstructing the depth of transparent regions by exploiting surrounding depth information, enhancing prediction accuracy in different backgrounds.
	
	\item We conduct extensive experiments on both datasets and real-world objects, demonstrating that our method substantially outperforms state-of-the-art approaches, validating its accuracy and generalization capabilities.
\end{enumerate}

The remainder of this paper is organized as follows. Section II reviews related work on transparent object depth prediction. Section III analyzes the TransCG dataset and current training strategies. Section IV presents the details of our proposed method. Section V describes extensive experiments conducted on both benchmark datasets and real-world scenarios to evaluate the proposed approach. Finally, Section VI concludes the paper and discusses future directions.


\section{Related Works}
Although many existing approaches for transparent object perception rely on multi-view information captured via camera motion \cite{albrecht2013seeing, zhou2019glassloc, dai2023graspnerf, klank2011transparent, ji2017fusing, kerr2022evo} or use specialized instruments for precise measurement \cite{han2015fixed, qian20163d}, such methods are often complex and difficult to deploy in real-world robotic systems. In contrast, single-view methods offer greater efficiency and practicality for broader robotic applications. As this paper focuses on single-view settings, we review prior work on transparent object perception, 6-DoF grasping, and monocular depth estimation within the single-view constraint. 

\subsection{Transparent Sensing}
We first introduce the datasets and describe how they preserve the original depth of transparent objects. Cleargrasp \cite{sajjan2020clear} is a pioneering work that introduced both real and synthetic datasets for transparent object reconstruction. It explicitly removed depth values in transparent regions using segmentation masks to avoid interference from unreliable sensor measurements. The OOD dataset \cite{zhu2021rgb} employs a similar strategy, removing depth values in transparent regions during training, and is frequently used in conjunction with ClearGrasp. TransAFF \cite{jiang2022a4t} also excludes depth information from transparent areas to mitigate the effects of sensor noise and improve training stability. However, the TODD dataset \cite{xu2021seeing} later demonstrated that completely ignoring depth in transparent regions yields suboptimal results. Similarly, the DREDS \cite{dai2022domain} and TransCG \cite{fang2022transcg} datasets advocate for preserving noisy depth in transparent areas, arguing that it can improve depth prediction performance.

\begin{figure*}[!t]\centering
	\includegraphics[width=6.8in]{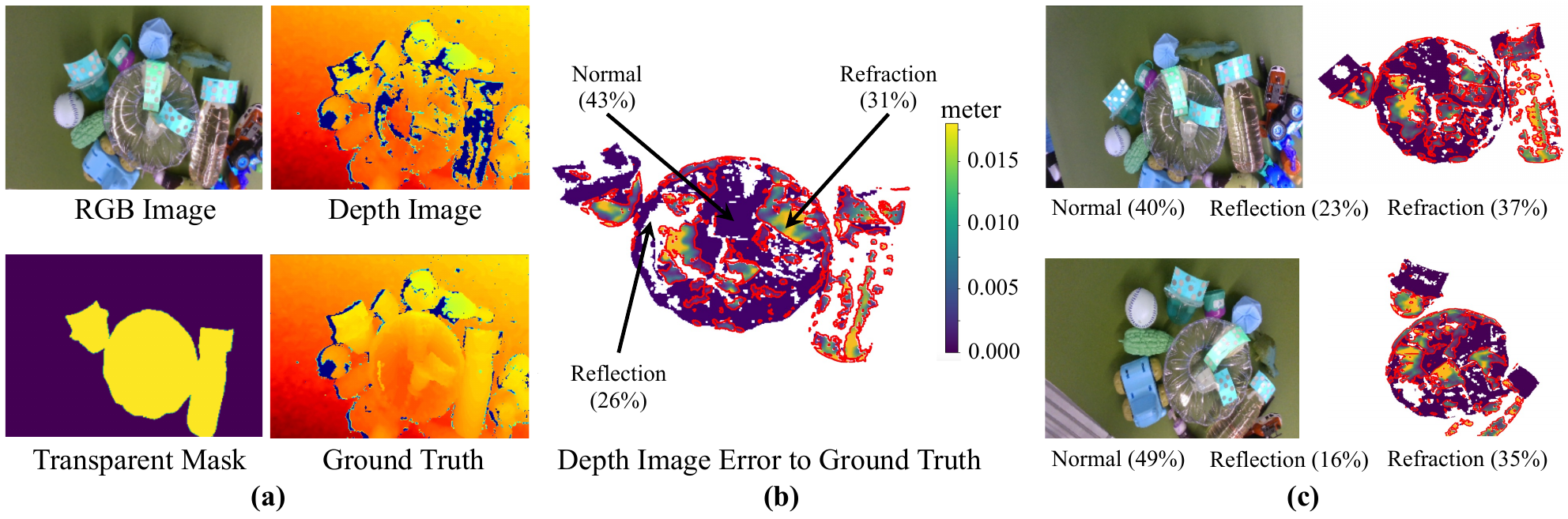}
	\caption{Visualization of the TransCG dataset \cite{fang2022transcg}. (a) Overview of the dataset structure. (b) Depth differences between the original and ground-truth images are analyzed across different transparent regions. Reflection areas are excluded due to missing depth. Refraction regions, while containing valid depth, exhibit significant prediction errors. In contrast, normal regions show no deviation from the ground truth. (c) The spatial distribution and proportion of these regions vary irregularly across different viewpoints.}\label{fig_2}
\end{figure*}

Despite the ongoing debate over whether to retain or discard depth in transparent regions, existing datasets typically employ full RGB-D images \cite{sajjan2020clear, zhu2021rgb, xu2021seeing, jiang2022a4t, fang2022transcg} or a combination of RGB and IR images \cite{dai2022domain}. These datasets all aim to reconstruct complete depth maps. Although these methods calculate full depth maps, they compute the loss only within transparent regions, encouraging the model to specifically focus on transparent depth prediction. Subsequent approaches based on these datasets typically emphasize learning correlations between RGB information and depth. They leverage cues such as boundaries and surface normals \cite{sajjan2020clear, fan2024tdcnet, tang2021depthgrasp}, or utilize affordance maps \cite{jiang2022a4t} to guide depth recovery. Several methods emphasize architectural innovations. These include UNet-based models such as DFNet \cite{fang2022transcg}, DepthGrasp \cite{tang2021depthgrasp}, and FDCT \cite{li2023fdct}; pyramid architectures like TCRNet \cite{zhai2024tcrnet}; and transformer-based designs such as SwinDRNet, which employs Swin Transformer encoders \cite{liu2021swin, dai2022domain}. Additionally, refinement network modules are used in OAGF \cite{yan2024transparent} and TDCNet \cite{fan2024tdcnet}. In contrast, TranspareNet \cite{xu2021seeing} introduces a different approach: it converts the depth map into a point cloud for feature extraction, then projects it back into depth space for final prediction.

In this work, we argue for retaining depth in transparent regions and provide experimental evidence supporting its effectiveness. Furthermore, we introduce a novel cue, relative depth, which enables the network to infer transparent object depth by learning its depth relationship with surrounding backgrounds. To support this, we compute the loss over the entire depth map rather than restricting it to transparent regions alone.

\subsection{Monocular Depth Estimation}
Studies have shown that humans estimate depth by observing the spatial relationships between objects \cite{oliva2007role}, which motivates us to explore similar cues for enhancing transparent object depth estimation. Through our observation, we find that monocular depth estimation can convert RGB images into non-measurement depth maps that encode relative depth information among objects.  In this work, we refer to these as relative depth maps.

Early monocular methods \cite{eigen2014depth, fu2018deep} often focused on improving metric accuracy within constrained domains, such as the NYU \cite{silberman2012indoor} and KITTI \cite{geiger2013vision} datasets. However, these approaches typically suffer from poor generalization when applied to novel scenes. Recent research has shifted toward zero-shot depth estimation, aiming to make reasonable depth predictions without relying on ground-truth annotations. To improve generalization, some methods treat depth estimation as a denoising task using diffusion models \cite{rombach2022high}, while others scale up training data to enhance robustness \cite{ranftl2020towards, yin2023metric3d}. DepthAnything \cite{yang2024depth} identifies serious deficiencies in existing real-world depth datasets and instead leverages synthetic images to train models using pseudo-labels. This strategy enables broad generalization across diverse object categories and scenes.

In this work, we do not propose a new monocular depth estimation model. Instead, we adopt the generalizable DepthAnything \cite{yang2024depth} model to generate relative depth maps. These maps which are used as auxiliary inputs during training to guide our network in learning depth relationships between transparent regions and their surroundings, ultimately improve the depth estimation accuracy.

\subsection{Robotic Grasping}
Grasping is a fundamental step in robotic interaction with the environment, where the goal is to generate feasible grasp poses based on the geometry of the target object. Mainstream approaches can be categorized into 4 degree-of-freedom (DoF) and 6-DoF grasping. 4-DoF methods \cite{lenz2015deep, wu2021real, shang2020deep, dong2024robotic, laili2022custom} typically generate grasp rectangles from images, constraining the grasp orientation to be aligned with the camera view direction, thus limiting the solution space. With advances in point cloud processing \cite{qi2017pointnet, qi2017pointnet++}, 6-DoF grasping, where grasp poses are generated directly in 3D space, has gained popularity due to its greater flexibility and expanded search space \cite{liang2019pointnetgpd, fang2020graspnet, sundermeyer2021contact, mousavian20196, wang2021graspness, cheng2025pcf}. To enable object-specific grasp generation, methods such as ContactGraspNet \cite{sundermeyer2021contact} and PCF-Grasp \cite{cheng2025pcf} apply instance masks to extract the depth of the target object before grasp prediction. This practice supports the rationale behind incorporating instance masks in our transparent object grasping framework.

Since single-view perception only captures the visible surface geometry from the camera's viewpoint, making accurate grasp generation is inherently challenging. This limitation becomes even more critical for transparent objects, where depth measurements are often noisy or missing. Therefore, recovering accurate depth maps of transparent objects is essential for enabling reliable 6-DoF grasp planning.

\section{Analysis of Transparent Completion}
\label{analysis_section}

\subsection{Preliminary}

In this paper, we focus on single-view transparent object depth completion, where the input consists of a monocular RGB image $I_i \in \mathbb{R}^{H \times W \times 3}$ and its corresponding depth image $D_i \in \mathbb{R}^{H \times W}$, and the goal is to predict a complete depth map $D_o \in \mathbb{R}^{H \times W}$ for downstream 6-DoF robotic grasping.

To facilitate depth prediction for transparent objects, several works \cite{fang2022transcg, sajjan2020clear, xu2021seeing} have constructed large-scale datasets containing RGB-D images $(I_i, D_i)$ captured from real or simulated environments, along with ground-truth depth maps $D_{gt} \in \mathbb{R}^{H \times W}$ and binary masks of transparent regions $M_{trans} \in \mathbb{R}^{H \times W}$, as illustrated in Fig. \ref{fig_2} (a).

\begin{figure*}[!t]\centering
	\includegraphics[width=6.8in]{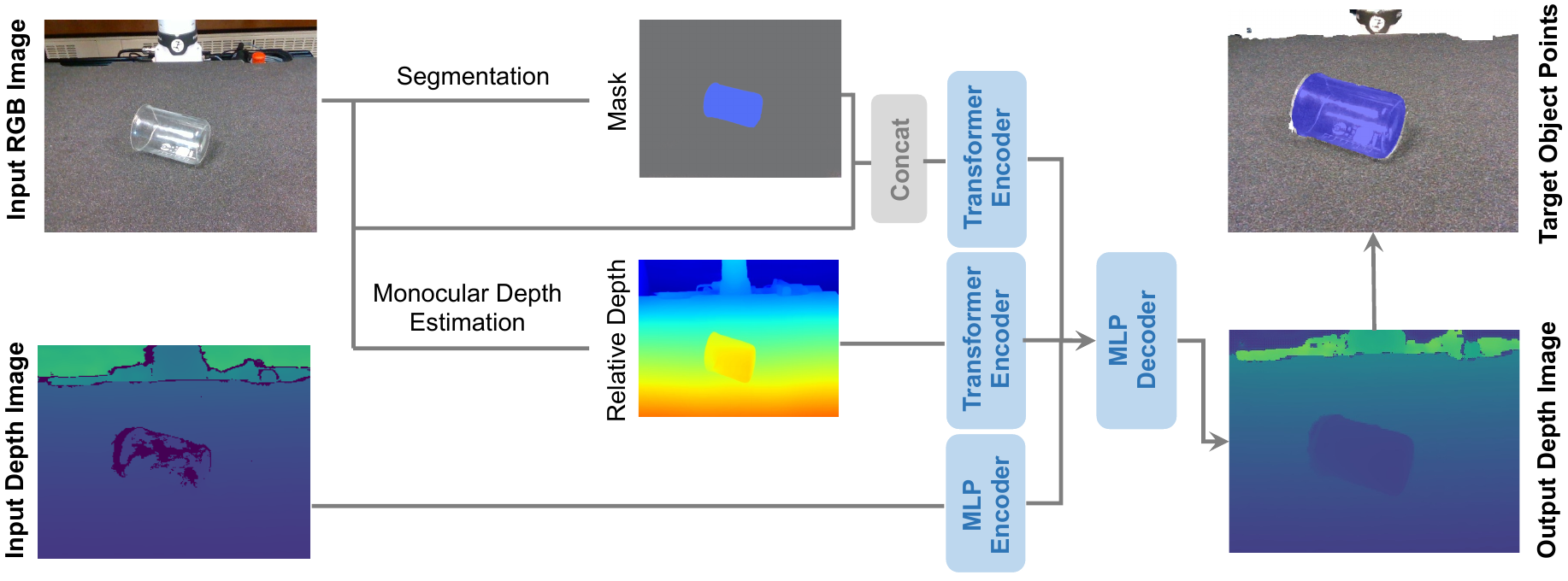}
	\caption{Overview of the ReMake framework. The instance mask and relative depth map are generated through segmentation and monocular depth estimation, respectively. The RGB image concatenated with the mask is encoded using a Transformer. The relative depth and original depth are encoded separately. All encoded features are then fused and decoded to predict the complete depth map. The target object's point cloud is then extracted using the instance mask.}\label{fig_3}
\end{figure*}

Some methods \cite{sajjan2020clear, zhu2021rgb, jiang2022a4t} regard the depth values in transparent regions as noisy and unreliable. These values are thus excluded using a transparency mask $M_{trans}$, resulting in a masked depth input $D_{imd}$. These methods then perform depth completion by feeding both the RGB image and the masked depth input into the network, as illustrated below:

\begin{align}
\label{formular_1}
	D_o = f(I_i, D_{imd})
\end{align}

However, recent works \cite{fang2022transcg, xu2021seeing, fan2024tdcnet} have found that retaining the full RGB-D input often leads to better performance. As a result, most current approaches input the unmodified RGB and depth images directly into the network, as shown below:
\begin{align}
\label{formular_2}
	D_o = f(I_i, D_i)
\end{align}

Since non-transparent regions already contain valid depth, their output can be directly retained. As a result, most methods compute the loss only over transparent regions using a masked loss, as defined below:
\begin{align}
\label{formular_3}
	L((D_o, D_{gt})|M_{trans})
\end{align}

This masked loss formulation enables the model to focus on reconstructing depth in transparent regions while excluding areas with already valid depth values.

\subsection{Dataset Analysis}
\label{dataset_analysis_section}

Although current methods achieve good performance on existing benchmarks, they often fail to generalize effectively to unseen scenes. This raises the question: why is their generalization poor? We hypothesize that these methods struggle to distinguish between valid and invalid depth values. To investigate this, we analyze and visualize the TransCG dataset \cite{fang2022transcg}.

Transparent sensing can be categorized into three distinct conditions: 1) Reflection: No reflected light returns to the camera, resulting in missing depth information. These regions are typically assigned a depth value of zero; 2) Refraction: Light passes through a transparent object and returns to the camera along a distorted path due to refraction, producing inaccurate depth readings; 3) Normal: Light returns along a correct path, producing accurate depth measurements. As shown in Fig. \ref{fig_2} (b), we visualize the depth error between input depth and ground-truth depth maps in TransCG \cite{fang2022transcg}.

Notably, the refraction and normal regions are fundamentally different. Although both provide non-zero depth values, only the normal regions yield accurate measurements, whereas refraction results in distorted and unreliable depth. Therefore, completely removing depth values in transparent regions also discards valid information, undermining the useful cues that could otherwise aid in depth prediction.

Moreover, the spatial distribution of these three regions can vary significantly under different viewpoints, as shown in Fig. \ref{fig_2} (c). As a result, methods that directly input RGB-D images into the network require the model to implicitly infer the reliability of depth values throughout the image. However, due to the complex and variable behavior of light in real-world scenes, such models often fail to generalize effectively beyond their training data.

This analysis suggests that explicitly guiding the model to identify transparent regions is essential for improving generalization to unseen environments, rather than relying on implicit learning alone. To this end, we propose ReMake, a novel framework that introduces a segmentation mask to clearly distinguish transparent areas, enabling the model to focus on reconstructing unreliable depth more effectively. Additionally, by introducing monocular depth estimation, the model is guided to capture the relative depth context between the transparent regions and their surrounding regions, offering broader spatial understanding than approaches that focus exclusively on transparent regions.

\section{Method}
\label{method_section}

\subsection{Method Overview}

In this section, we will introduce the detailed architecture of the proposed ReMake, as shown in Fig. \ref{fig_3}. Our framework consists of two key components: mask-attention encoding and the relative depth cue module.
We first concatenate the RGB image with the transparent region mask $M_{trans}$, and feed it into an encoder to output $\mathcal{F}_{mask}$, which enhances the distinction between the transparent object and the background (Sec. \ref{mask_section}). 
Meanwhile, the relative depth map $D_{rel} \in \mathbb{R}^{H \times W}$ generated from monocular depth estimation is input into another encoder to extract the depth relationship feature $\mathcal{F}_{rel}$ between the transparent region and its surrounding areas, thereby providing a localization of the transparent object (Sec. \ref{relative_depth_section}).
The depth image is solely input to an MLP encoder to extract the depth feature $\mathcal{F}_{depth}$. Finally, $\mathcal{F}_{mask}$, $\mathcal{F}_{rel}$ and $\mathcal{F}_{depth}$ are combined into a residual structure decoder to predict the complete depth map $D_o$. The overall framework is formulated as:

\begin{align}
\label{formular_4}
	D_o = f(I_i, M_{trans}, D_{rel}, D_i)
\end{align}

\subsection{Mask Attention}
\label{mask_section}

As shown in Sec. \ref{dataset_analysis_section}, the depth of transparent regions is complex, yet it still contains valuable depth cues that can help depth reconstruction. To reduce the model's learning burden while retaining valuable depth cues in transparent regions, we introduce a mask that explicitly separates these regions from the background. This enables the model to concentrate on reconstructing depth within the masked areas by effectively leveraging both RGB and depth inputs.

To implement this, given an RGB image $I_i$ and depth image $D_i$, we first concatenate the transparent region mask $M_{trans}$ with the RGB image to form the mask feature $\mathcal{F}_{mask} \in \mathbb{R}^{H \times W \times 4}$, which is passed through a Swin-Transformer \cite{liu2021swin} encoder for multi-layer feature extraction, following the design of TDCNet \cite{fan2024tdcnet}.

\subsection{Relative Depth}
\label{relative_depth_section}

Since transparent objects exhibit relative spatial relationships with nearby objects, we propose leveraging relative depth information to guide the reconstruction of their depth more accurately. We observe that monocular depth estimation \cite{eigen2014depth, fu2018deep, silberman2012indoor, geiger2013vision, rombach2022high, ranftl2020towards, yin2023metric3d} provides non-metric depth maps that represent relative depth relationships. Notably, these depth maps lack real-world metrics and cannot be directly used for grasping tasks. However, they still effectively capture spatial relationships, which are particularly valuable for estimating the depth of transparent regions.

As shown in Fig. \ref{fig_3}, based on the input RGB image, we use DepthAnything \cite{yang2024depth} to generate a relative depth image $D_{rel}$. This map is then passed through a Swin Transformer encoder \cite{liu2021swin}, which extracts multilayer features $\mathcal{F}_{rel}$ that are spatially aligned with the RGB and mask features. Notably, to preserve the global information of the relative depth map, we avoid concatenating it with the mask. Instead, it is processed independently through a separate encoder branch.

\subsection{Loss Function}
\label{loss_section}
Since our method infers transparent depth based on spatial relationships with surrounding objects, globally accurate depth prediction is crucial. Therefore, unlike previous methods that apply masked loss only within transparent regions (Eq. \ref{formular_3}), we adopt a global $L1$ loss computed over all pixels: $L(D_o, D_{gt})$. We refer to these two types of supervision as mask loss and global loss in the following sections.

\subsection{Implementation Details}

{\bf{Training:}} 
We applied the TransCG dataset \cite{fang2022transcg} for training, evaluation, and real-world experiments. The model was trained with global loss by default. For comparative analysis, we additionally trained the model using mask loss, following previous methods \cite{fan2024tdcnet, fang2022transcg, liu2024transparent}, and used it for evaluation and visualization purposes. 

{\bf{Metrics:}}
For depth estimation on the TransCG dataset, we choose the root mean square error (RMSE), the absolute relative difference (REL), the mean absolute error (MAE), and the threshold $\delta$ as metrics. It is important to note that results obtained on the dataset do not always align with real-world performance, reflecting the challenges of generalization.

\begin{table*}[!ht]\centering
    \caption{Quantitative Comparison on the TransCG Dataset.}
    \label{table_1}
    \begin{tabular}{ccccccccc}
    \toprule
        {\textbf{Methods}} & {RMSE$\downarrow$} & {REL$\downarrow$} & {MAE$\downarrow$} & {$\delta 1.01 \uparrow$} & {$\delta 1.03 \uparrow$} & {$\delta 1.05 \uparrow$} & {$\delta 1.10 \uparrow$} & {$\delta 1.25 \uparrow$} \\
    \midrule
    		{CG \cite{sajjan2020clear}} & 0.054 & 0.083 & 0.037 & - & - & 50.48 & 68.68 & 95.29 \\
    		{DFNet \cite{fang2022transcg}} & 0.018 & 0.027 & 0.012 & - & - & 81.15 & 95.80 & 99.87 \\
    		{LIDF-Refine \cite{zhu2021rgb}} & 0.019 & 0.034 & 0.015 & - & - & 78.22 & 94.26 & 99.80 \\
    		{TCRNet \cite{zhai2024tcrnet}} & 0.017 & 0.020 & 0.010 & - & - & 88.96 & 96.94 & 99.87\\
    		{TransparentNet \cite{xu2021seeing}} & 0.026 & 0.023 & 0.013 & - & - & 88.45 & 96.25 & 99.42 \\
    		{FDCT \cite{li2023fdct}} & 0.015 & 0.022 & 0.010 & - & - & 88.18 & 97.15 & 99.81 \\
    		{TODE-Trans \cite{chen2023tode}} & 0.013 & 0.019 & \underline{0.008} & - & - & 90.43 & 97.39 & 99.81 \\
    		{DualTransNet \cite{liu2024transparent}} & 0.012 & 0.018 & \underline{0.008} & - & - & 92.37 & 97.98 & 99.81 \\
    		{TDCNet (Global Loss) \cite{fan2024tdcnet}} & 0.023 & 0.034 & 0.014 & 24.77 & 61.19 & 79.88 & 94.64 & 99.87 \\
    		{TDCNet (Mask Loss) \cite{fan2024tdcnet}} & 0.012 & \underline{0.017} & \underline{0.008} & \underline{54.18} & 82.17 & 92.25 & 97.66 & 99.87 \\
     \midrule
        \textbf{ReMake (Global Loss)} & \underline{0.011} & \underline{0.017} & \underline{0.008} & 47.35 & \underline{85.12} & \underline{93.91} & \underline{98.67} & \textbf{99.97} \\
        \textbf{ReMake (Mask Loss)} & \textbf{0.010} & \textbf{0.015} & \textbf{0.007} & \textbf{54.20} & \textbf{88.89} & \textbf{94.76} & \textbf{98.73} & \underline{99.96} \\
    \bottomrule
    \end{tabular}
	\begin{tablenotes}
		\item \textbf{Bold} refers to the highest results. \underline{Underline} refers to the second-highest results.
	\end{tablenotes}

\end{table*}

\begin{table*}[h]
\centering
\caption{Grasp Success Rates under Each View.}
\label{table_2}
\renewcommand\cellalign{cc}
\begin{tabular}{ccccccccccc}
\toprule
\textbf{View} & \textbf{Method} & tube(L) & tube(S) & tube(L/w) & bottle & beaker & wine glass & cup & mean \\
\midrule
\multirow{3}{*}{Top-down} 
& DFNet \cite{fang2022transcg} & 4/7 & 3/7 & 3/7 & 4/7 & 3/7 & 1/7 & 3/7 & 42.86\% \\
& TDCNet \cite{fan2024tdcnet} & 5/7 & 5/7 & 3/7 & 5/7 & 4/7 & 4/7 & 3/7 & 59.18\% \\
& \textbf{ReMake} & \textbf{6/7} & \textbf{6/7} & \textbf{6/7} & \textbf{7/7} & \textbf{6/7} & \textbf{5/7} & \textbf{6/7} & \textbf{85.71\%} \\
\midrule
\multirow{3}{*}{Bird-eye} 
& DFNet \cite{fang2022transcg} & 2/7 & 0/7 & 1/7 & 1/7 & 0/7 & 0/7 & 1/7 & 10.20\% \\
& TDCNet \cite{fan2024tdcnet} & 3/7 & 2/7 & 1/7 & 0/7 & 1/7 & 2/7 & 1/7 & 20.41\% \\
& \textbf{ReMake} & \textbf{5/7} & \textbf{4/7} & \textbf{5/7} & \textbf{6/7} & \textbf{5/7} & \textbf{6/7} & \textbf{5/7} & \textbf{73.47\%} \\
\midrule
\multirow{3}{*}{Horizontal} 
& DFNet \cite{fang2022transcg} & 0/7 & 0/7 & 0/7 & 1/7 & 0/7 & 0/7 & 0/7 & 2.04\% \\
& TDCNet \cite{fan2024tdcnet} & 0/7 & 1/7 & 1/7 & 1/7 & 1/7 & 0/7 & 1/7 & 8.16\% \\
& \textbf{ReMake} & \textbf{6/7} & \textbf{4/7} & \textbf{4/7} & \textbf{5/7} & \textbf{6/7} & \textbf{5/7} & \textbf{5/7} & \textbf{71.43\%} \\
\bottomrule
\end{tabular}
\end{table*}

\begin{table}[htb]
\centering
\caption{Comparison of Accuracy Across Different Regions on the TransCG Dataset.}
\label{table_3}
\begin{tabular}{cccccc}
\toprule
\textbf{Region} & \textbf{Metric} & \makecell{\textbf{ReMake} \\ \textbf{(Global)}} & \makecell{\textbf{ReMake} \\ \textbf{(Mask)}} & \makecell{\textbf{TDCNet \cite{fan2024tdcnet}} \\ \textbf{(Mask)}} & \makecell{\textbf{Dataset} \\ \textbf{Ratio} } \\
\midrule
\multirow{2}{*}{Refraction} & RMSE$\downarrow$ & 0.0113 & \textbf{0.0103} & 0.0113 & \multirow{2}{*}{60.08\%} \\
                            & MAE$\downarrow$  & 0.0082 & \textbf{0.0072} & 0.0080  \\
\midrule
\multirow{2}{*}{Reflection} & RMSE$\downarrow$ & 0.0207 & 0.0194 & \textbf{0.0191} & \multirow{2}{*}{17.47\%} \\
                            & MAE$\downarrow$  & 0.0166 & 0.0149 & \textbf{0.0144}  \\
\midrule
\multirow{2}{*}{Normal}     & RMSE$\downarrow$ & 0.0108 & 0.0076 & \textbf{0.0066} & \multirow{2}{*}{22.45\%} \\
                            & MAE$\downarrow$  & 0.0090 & 0.0061 & \textbf{0.0031} \\
\bottomrule
\end{tabular}
\end{table}

\begin{itemize}
	\item RMSE: $\sqrt{\frac{1}{|\widehat{D}|} \sum_{d \in \widehat{D}}\left\|d-d^*\right\|^2}$
	\item REL: $\frac{1}{|\widehat{D}|} \sum_{d \in \widehat{D}} |\frac{d-d^*}{d*}|$
	\item MAE: $\frac{1}{|\widehat{D}|} \sum_{d \in \widehat{D}}\left|d-d^*\right|$
	\item threshold $\delta$: $\left(\frac{d_i}{d_i^*}, \frac{d_i^*}{d_i}\right)<\delta$, where the $\delta$ is set to 1.01, 1.03, 1.05, 1.10, and 1.25. Notably, we introduce the stricter thresholds of 1.01 and 1.03, which are not included in previous works \cite{sajjan2020clear, fang2022transcg, fan2024tdcnet}, to better reflect the high precision required for grasping tasks. Since our method generates completed point clouds for grasp execution, and point cloud accuracy directly influences grasp success, these tighter thresholds are essential for realistic evaluation. Specifically, when the camera is positioned $50 \ cm$ from the transparent object, a threshold of 1.01 corresponds to a point cloud error within $\pm 5 \ mm$, which is essential to ensure reliable grasping with the 8 cm wide gripper of the Franka robotic arm. 
\end{itemize}

{\bf{Model Implementation:}} We use an Intel Xeon 6226R CPU and eight NVIDIA GeForce RTX 3090 GPUs for training, with a single GPU used for testing. Real-world experiments are conducted on a computer equipped with an AMD 7700X CPU and a single NVIDIA GeForce RTX 3090 GPU. We set the batch size to 8 and the input image size to 640 $\times$ 480. We employ the AdamW optimizer with an initial learning rate of 0.001, which is decayed to one-tenth every 15 epochs over a total of 40 epochs. The Transformer and MLP encoders follow the design of TDCNet \cite{fan2024tdcnet}, while the decoder is implemented as a multi-layer residual MLP, inspired by ResNet \cite{he2016deep}. Since our method requires an instance mask as input, we use Segment Anything \cite{kirillov2023segment} for mask generation in real-world experiments. It is worth noting that any segmentation algorithm can be used; we select Segment Anything because it can segment transparent objects in diverse scenes without retraining. However, it is relatively time-consuming, taking approximately 2.15 seconds per frame on our real-world setup. The total processing time per frame is around 2.16 seconds, with monocular depth estimation requiring 0.0032 seconds and depth map prediction taking 0.0105 seconds.

\section{Experiment}
\label{experiment_section}

In this section, we evaluate the proposed method on both benchmark datasets and real-world grasping experiments, comparing its performance against state-of-the-art methods. Finally, we conduct ablation studies to evaluate the contribution of each module within the proposed framework.

\subsection{Dataset Completion Results}
\label{dataset_section}

We compare our method with state-of-the-art approaches on the TransCG dataset, and the quantitative results are reported in Table \ref{table_1}. Our model achieves the best overall performance across evaluation metrics. When trained using loss computed solely within the masked transparent regions, it obtains the highest accuracy on the dataset. Using a global loss results in a slight reduction in performance, though it remains competitive. In contrast, retraining TDCNet \cite{fan2024tdcnet} with a global loss causes a notable degradation in accuracy, indicating that our framework is more robust to supervision strategy changes.

\begin{figure*}[!htb]\centering
	\includegraphics[width=6.9in]{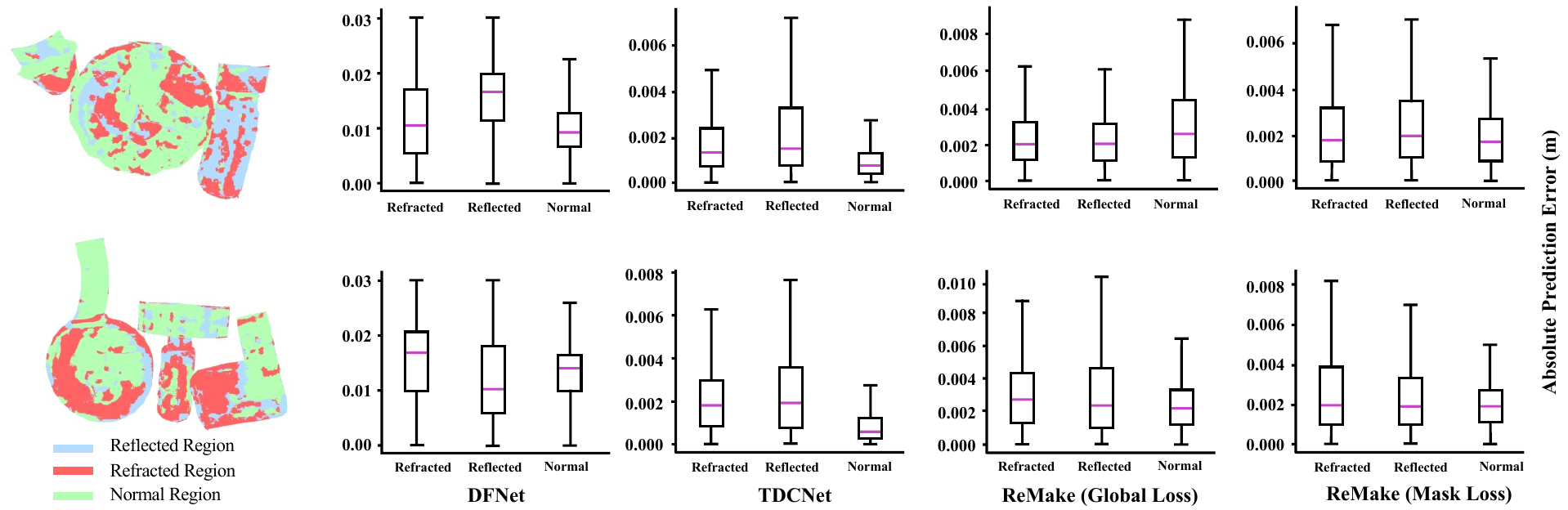}
	\caption{Boxplot comparison of per-pixel absolute depth prediction errors across three region types (Refracted, Reflected, and Normal) on the TransCG dataset.}\label{fig_5}
\end{figure*}

\begin{figure*}[!htb]\centering
	\includegraphics[width=6.9in]{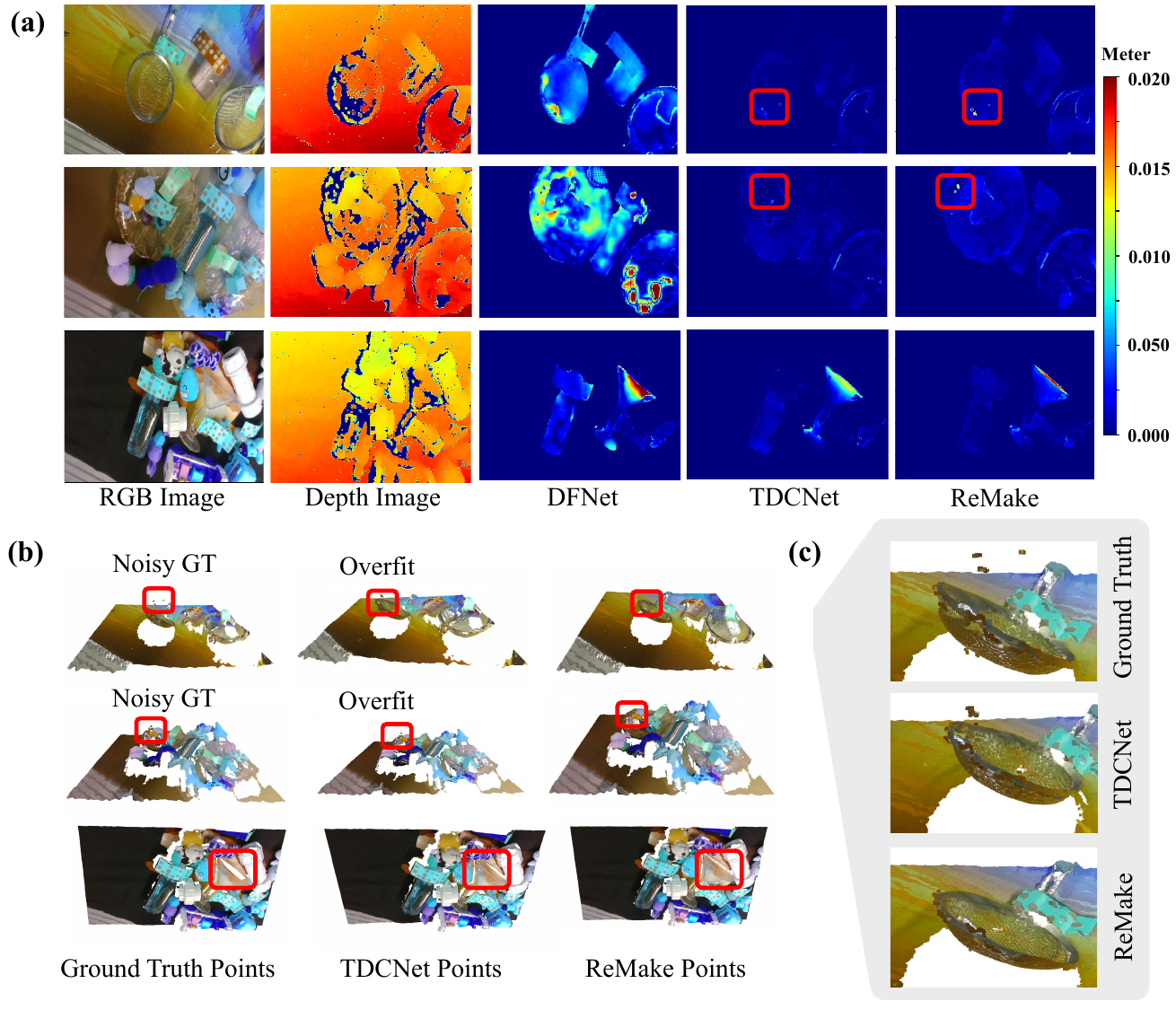}
	\caption{Qualitative comparisons on the TransCG dataset. (a) Input RGB-D image and the corresponding depth prediction error map relative to the ground truth. In regions with noisy ground-truth depth (highlighted by red boxes), TDCNet exhibits overfitting, whereas ReMake demonstrates greater robustness and produces cleaner, more consistent point cloud reconstructions. (b) Point clouds generated from the predicted depth maps. (c) Magnified view of the point cloud from the first row for detailed visual comparison.}\label{fig_4}
\end{figure*}

\begin{figure*}[!htb]\centering
	\includegraphics[width=6.8in]{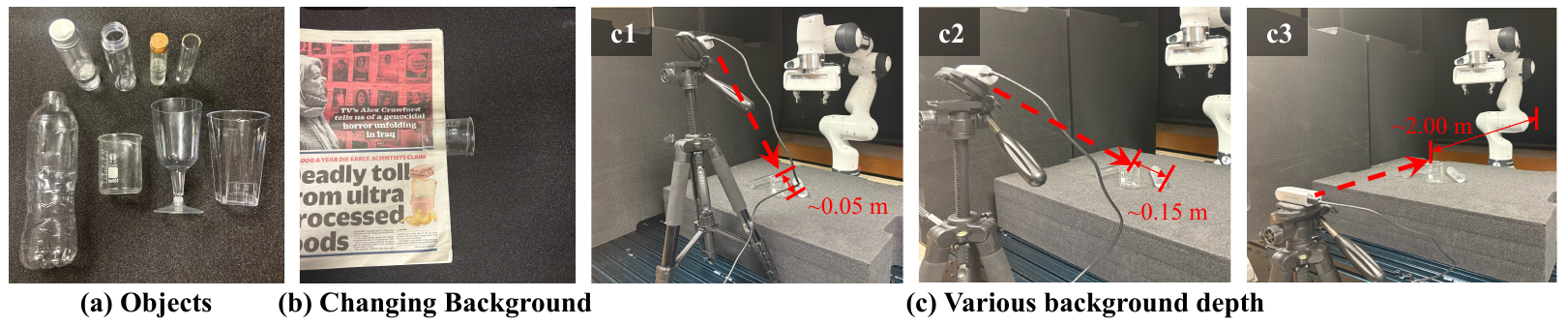}
	\caption{Real-world experimental setup. (a) Transparent objects for grasping. (b) Background variations. (c1-c3) Top-down $(0.05m)$, bird-eye $(0.15m)$, and horizontal $(2.00m)$ views,  illustrating varying background depths that influence depth prediction performance.}\label{fig_6}
\end{figure*}

\begin{figure*}[!htb]\centering
	\includegraphics[width=6.8in]{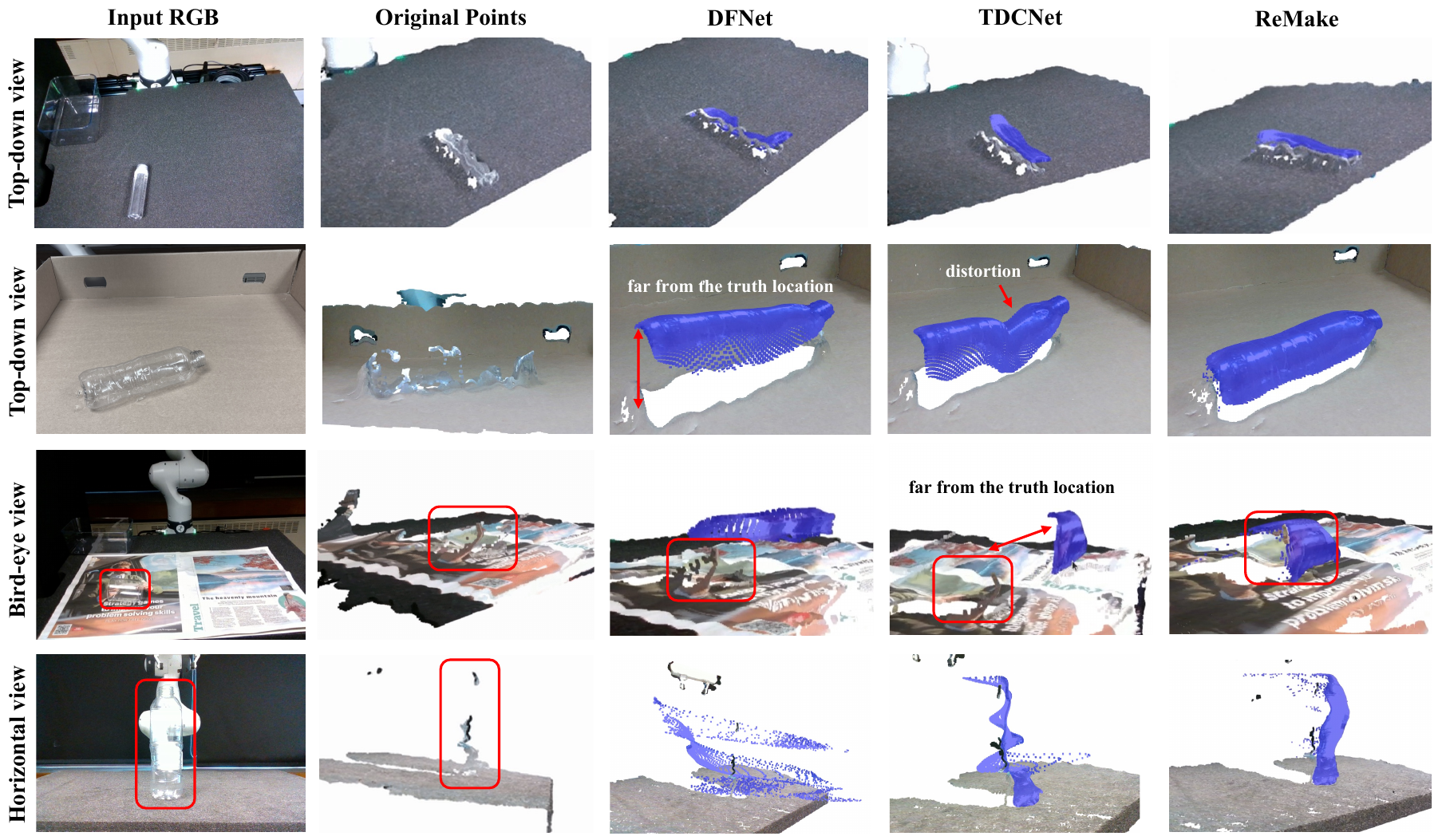}
	\caption{Point cloud visualizations from real-world experiments across multiple camera viewpoints. Due to the top-down capture bias of the TransCG dataset \cite{fang2022transcg}, models such as DFNet \cite{fang2022transcg} and TDCNet \cite{fan2024tdcnet} exhibit poor generalization to novel views. In comparison, ReMake consistently demonstrates robust performance regardless of camera orientation.}\label{fig_7}
\end{figure*}

To better understand model behavior across different transparent region types, we analyze prediction errors in various regions. Fig. \ref{fig_5} visualizes the per-pixel error and the average error for two representative examples. As shown, DFNet \cite{fang2022transcg} exhibits the highest overall error. TDCNet performs well in normal regions, but its error increases notably in refractive and reflective areas. In comparison, our method achieves more balanced errors across all three region types.

We further report the average error of TDCNet across the entire dataset in Table \ref{table_3}. While TDCNet with masked loss achieves the lowest error in normal regions, its performance degrades significantly in refractive areas. This suggests that the model over-relies on the original depth input, reducing its robustness in regions where depth is distorted. In contrast, our method maintains more balanced errors across all region types, indicating that it learns to treat all masked regions as potentially unreliable and consistently reconstructs depth using both RGB-D inputs and contextual cues.

Comparing our global and masked loss settings, we observe that restricting supervision to masked regions slightly improves performance on the dataset. However, we adopt the global loss as our default, since synthetic dataset performance does not always correlate with real-world results. This decision is further demonstrated in our real-world ablation experiments (see Sec. \ref{ablation_section}).

In Fig. \ref{fig_4}, we show qualitative comparisons between our method and other models on the TransCG dataset. It is worth noting that the ground-truth depth maps contain certain inaccuracies. While TDCNet \cite{fan2024tdcnet} tends to overfit to these imperfect labels. In contrast, our model, informed by relative depth cues, exhibits improved robustness and better maintains the true geometric structure of the scene.

\begin{figure*}[!t]\centering
	\includegraphics[width=6.8in]{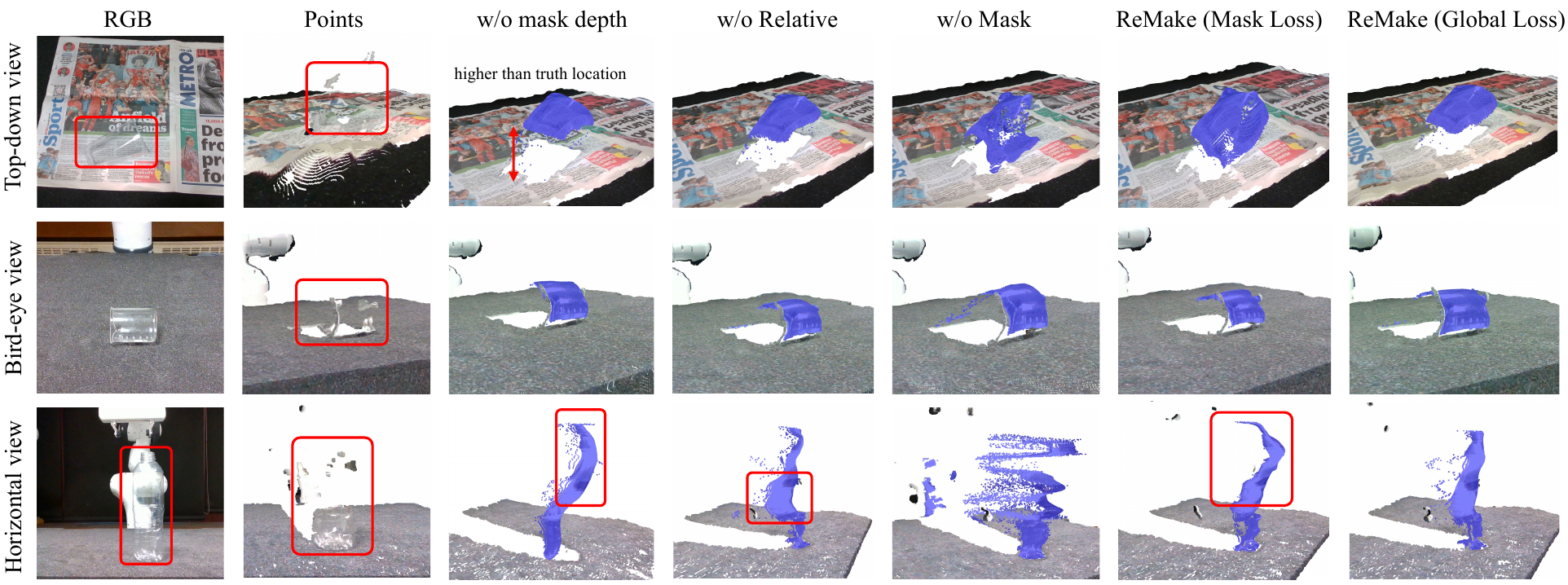}
	\caption{Point cloud visualizations from real-world ablation studies under different camera views.}\label{fig_9}
\end{figure*}

\begin{table*}[!ht]\centering
    \caption{Quantitative Ablation Results on the TransCG Dataset.}
    \label{table_4}
    \begin{tabular}{ccccccccc}
    \toprule
        {\textbf{Methods}} & {RMSE$\downarrow$} & {REL$\downarrow$} & {MAE$\downarrow$} & {$\delta 1.01 \uparrow$} & {$\delta 1.03 \uparrow$} & {$\delta 1.05 \uparrow$} & {$\delta 1.10 \uparrow$} & {$\delta 1.25 \uparrow$} \\
    \midrule
    		{ReMake blank} & 0.016 & 0.023 & 0.010 & 43.13 & 79.32 & 88.99 & 96.22 & 99.67 \\
    		{ReMake w/o Trans Depth} & 0.012 & 0.020 & 0.009 & 41.50 & 81.76 & 92.06 & 97.93 & 99.94\\
    		{ReMake w/o Relative} & \textbf{0.010} & \textbf{0.016} & \textbf{0.007} & \textbf{50.89} & \textbf{86.40} & \textbf{94.26} & \underline{98.65} & \underline{99.96} \\
    		{ReMake w/o Mask} & 0.015 & 0.020 & 0.009 & 45.62 & 82.65 & 90.86 & 96.70 & 99.72 \\
    		
     \midrule
        \textbf{ReMake} & \underline{0.011} & \underline{0.017} & \underline{0.008} & \underline{47.35} & \underline{85.12} & \underline{93.91} & \textbf{98.67} & \textbf{99.97} \\
    \bottomrule
    \end{tabular}
	\begin{tablenotes}
		\item \textbf{Bold} refers to the highest results. \underline{Underline} refers to the second-highest results.
	\end{tablenotes}

\end{table*}

\subsection{Real-world Grasping Experiments}
\label{realworld_section}

\subsubsection{Experiment Setting}

To evaluate our method in real-world conditions, we use a Franka robot to conduct object grasping experiments. The grasped objects include: a bottle, a large tube (tube L, radius $32 mm$), a small tube tube S, radius $22 mm$), a tube filled with water (tube L/w) a beaker, a wine glass, and a cup, as shown in Fig. \ref{fig_6} (a). The small tube exhibits stronger refraction effects due to its higher curvature, while the presence of water in the large tube further alters its refraction behavior.

To assess the model's robustness under varying visual conditions, we test it with both background color variations and changes in background depth. As shown in Fig. \ref{fig_6} (c), three camera viewpoints are used: top-down, bird-eye, and horizontal, representing progressively increasing distances between the background and the object. In the top-down and bird-eye views, different background textures, including colored paper and newspapers, are randomly placed beneath the objects. The horizontal view naturally contains a complex and variable background due to its viewing angle.

We use PCF-Grasp\cite{cheng2025pcf} as the 6-DoF grasp generation algorithm to generate grasp proposals based on a single-view depth image. As the grasp algorithm needs a segmentation mask to choose the target object, we use the same mask as the model input. Specifically, we employ the Segment Anything model \cite{kirillov2023segment} to generate this mask. For each experiment, a single object is placed on the table, and grasp proposals are generated. Each object undergoes seven trials per camera view.

The quantitative grasping results are reported in Table \ref{table_2}. Since the TransCG dataset \cite{fang2022transcg} is collected from a top-down camera view, existing methods generally perform well under this setting, though some still exhibit minor distortions and inaccuracies. However, when the camera viewpoint shifts to a bird-eye or horizontal, the spatial relationship between the transparent object and the background becomes more complex, typically with a greater distance between them, which significantly increases the difficulty of accurate depth prediction, as Fig. \ref{fig_7}. As a result, the grasp success rate of previous methods \cite{fang2020graspnet, fan2024tdcnet} drops sharply under these new viewpoints. In contrast, our method not only achieves state-of-the-art performance in the top-down view but also maintains robust grasp success rates across both bird-eye and horizontal views. This demonstrates the model's ability to adapt to changing backgrounds and varying spatial configurations, thereby demonstrating stronger generalization capabilities in practical scenarios.

\subsection{Ablation Study}
\label{ablation_section}

To show the effectiveness of our modules, we evaluate our model with the following settings:

\begin{itemize}
	\item ReMake Blank: ours without mask and relative depth.
	\item ReMake w/o Rel: ours without relative depth.
	\item ReMake w/o Mask: ours without a mask.
	\item ReMake w/o mask depth: removing the mask depth before inputting to the model \cite{sajjan2020clear, zhu2021rgb}.
	
\end{itemize}

\begin{figure*}[!t]\centering
	\includegraphics[width=6.8in]{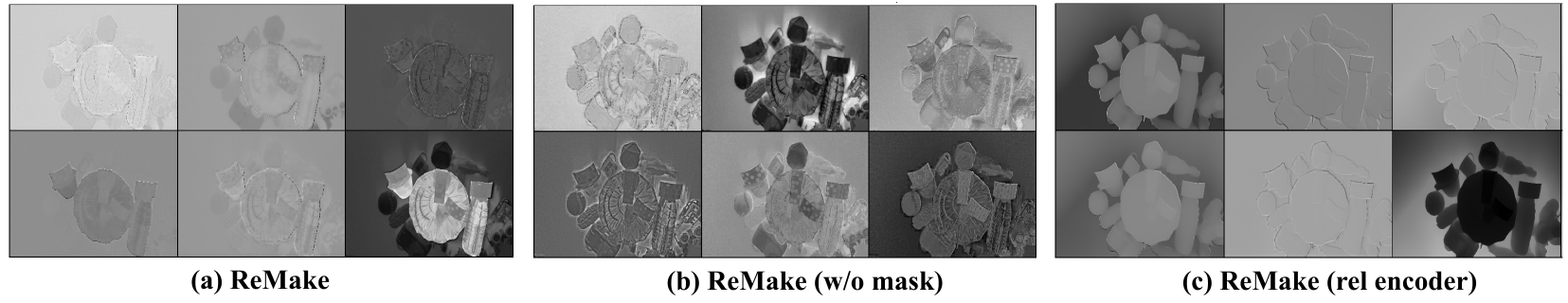}
	\caption{Encoder feature map visualization. (a) ReMake with a mask clearly highlights transparent regions. (b) Without a mask, the encoder fails to distinguish them from the background. (c) The relative depth encoder captures coarse object relations but loses fine-grained depth details.}\label{fig_8}
\end{figure*}

We conduct training and evaluation on the TransCG dataset \cite{fang2022transcg}, with results summarized in Table \ref{table_4}. We further validate our method through real-world experiments under various camera viewpoints, as shown in Fig. \ref{fig_9}. To provide a more complete evaluation, we analyze both the synthetic and real-world results in combination. 

Integrating the instance mask significantly improves prediction accuracy on the TransCG dataset, while models without the mask exhibit substantial performance degradation in real-world scenarios. As shown in Fig. \ref{fig_8}, incorporating the mask (Fig. \ref{fig_8} (a)) enables clear separation between transparent objects and the background. This mask-guided encoding directs the model's attention to ambiguous regions, helping it focus on reconstructing depth where it is most uncertain, thereby enhancing generalization to real-world scenarios.

Although incorporating relative depth alone contributes to performance improvements, combining it with the mask leads to a slight performance drop on the dataset. This is because relative depth lacks fine-grained geometry details and only provides coarse depth relationships between objects, as reflected in the encoder's feature maps (Fig. \ref{fig_8} c). As shown in Fig. \ref{fig_9}, relative depth helps maintain prediction accuracy despite dramatic changes in viewpoint and background, thereby enhancing generalization. Similarly, while using masked loss alone can achieve better accuracy on the dataset by narrowing the supervision region (see Table \ref{table_1}), it performs poorly in real-world tests (see Fig. \ref{fig_9}) due to its neglect of surrounding depth context.

We also evaluate the impact of removing depth values in transparent regions before inputting them into the network. Results indicate that this approach degrades performance on both the dataset and in real-world scenarios. This decline occurs because the removal process eliminates not only unreliable refracted and reflected depths but also valid depth information from normal regions, thereby reducing the model's effective input and impairing its predictive accuracy.

\section{Discussion and Conclusion}

In this work, we propose a novel framework for transparent object depth completion, guided by relative depth cues and instance masks. The mask explicitly distinguishes unreliable transparent regions from background areas, allowing the model to focus on reconstructing transparent object depth from RGB-D input during training. Additionally, by incorporating a relative depth map, the model learns spatial relationships between transparent and surrounding objects to improve depth prediction accuracy. Extensive experiments on both benchmark datasets and real-world scenarios demonstrate the strong generalization ability and robustness of our method across unseen scenes and varied camera viewpoints.

\
\noindent {\bf{Discussion and Limitations:}}

Our experiments show that removing depth information from transparent regions degrades prediction accuracy, which we attribute primarily to the loss of valid depth values originally present in the raw sensor data. This is further evidenced by reduced performance in scenes where transparent objects lack reliable depth entirely. However, it remains unclear whether retained refracted depths contribute positively to reconstruction. A more detailed analysis of these regions could inform improved depth completion strategies in future work.

Our model is primarily trained on the TransCG \cite{fang2022transcg} dataset, where the depth sensor was positioned $50-70 cm$ from the target object during data collection. As a result, the model performs best within this range. Extending the training dataset to include a wider range of camera-object distances may improve performance under more diverse settings.

Both the relative depth estimation model and the depth completion model are generative models and tend to produce closed surfaces. As a result, they struggle to represent internal hollow structures of transparent objects-e.g., beakers are often reconstructed as solid objects. In future work, we plan to explore representations that better preserve the internal geometry of transparent objects.

\bibliographystyle{IEEEtranTIE}
\bibliography{main}

\begin{thebibliography}{10}
\providecommand{\url}[1]{#1}
\csname url@samestyle\endcsname
\providecommand{\newblock}{\relax}
\providecommand{\bibinfo}[2]{#2}
\providecommand{\BIBentrySTDinterwordspacing}{\spaceskip=0pt\relax}
\providecommand{\BIBentryALTinterwordstretchfactor}{4}
\providecommand{\BIBentryALTinterwordspacing}{\spaceskip=\fontdimen2\font plus
\BIBentryALTinterwordstretchfactor\fontdimen3\font minus \fontdimen4\font\relax}
\providecommand{\BIBforeignlanguage}[2]{{%
\expandafter\ifx\csname l@#1\endcsname\relax
\typeout{** WARNING: IEEEtran.bst: No hyphenation pattern has been}%
\typeout{** loaded for the language `#1'. Using the pattern for}%
\typeout{** the default language instead.}%
\else
\language=\csname l@#1\endcsname
\fi
#2}}
\providecommand{\BIBdecl}{\relax}
\BIBdecl

\bibitem{klank2011transparent}
U.~Klank, D.~Carton, and M.~Beetz, ``Transparent object detection and reconstruction on a mobile platform,'' in \emph{IEEE Int. Conf. Rob. Autom.(ICRA)}, pp. 5971--5978, 2011.

\bibitem{dai2023graspnerf}
Q.~Dai, Y.~Zhu, Y.~Geng, C.~Ruan, J.~Zhang, and H.~Wang, ``Graspnerf: Multiview-based 6-dof grasp detection for transparent and specular objects using generalizable nerf,'' in \emph{IEEE Int. Conf. Rob. Autom.(ICRA)}, pp. 1757--1763, 2023.

\bibitem{kerr2022evo}
J.~Kerr, L.~Fu, H.~Huang, Y.~Avigal, M.~Tancik, J.~Ichnowski, A.~Kanazawa, and K.~Goldberg, ``Evo-nerf: Evolving nerf for sequential robot grasping of transparent objects,'' in \emph{Conf. on Rob. Learn.(CoRL)}, 2022.

\bibitem{zhou2019glassloc}
Z.~Zhou, T.~Pan, S.~Wu, H.~Chang, and O.~C. Jenkins, ``Glassloc: plenoptic grasp pose detection in transparent clutter,'' in \emph{IEEE Int. Conf. Intell. Rob. Syst.(IROS)}, pp. 4776--4783, 2019.

\bibitem{albrecht2013seeing}
S.~Albrecht and S.~Marsland, ``Seeing the unseen: Simple reconstruction of transparent objects from point cloud data,'' in \emph{Robot.: Sci. Syst. (RSS)}, vol.~3, pp. 1--6, 2013.

\bibitem{ji2017fusing}
Y.~Ji, Q.~Xia, and Z.~Zhang, ``Fusing depth and silhouette for scanning transparent object with rgb-d sensor,'' \emph{Int. J. Opt.}, vol. 2017, no.~1, 2017.

\bibitem{sajjan2020clear}
S.~Sajjan, M.~Moore, M.~Pan, G.~Nagaraja, J.~Lee, A.~Zeng, and S.~Song, ``Clear grasp: 3d shape estimation of transparent objects for manipulation,'' in \emph{IEEE Int. Conf. Rob. Autom.(ICRA)}, pp. 3634--3642, 2020.

\bibitem{fang2022transcg}
H.~Fang, H.-S. Fang, S.~Xu, and C.~Lu, ``Transcg: A large-scale real-world dataset for transparent object depth completion and a grasping baseline,'' \emph{IEEE Rob. Auto. Lett.}, vol.~7, no.~3, pp. 7383--7390, 2022.

\bibitem{dai2022domain}
Q.~Dai, J.~Zhang, Q.~Li, T.~Wu, H.~Dong, Z.~Liu, P.~Tan, and H.~Wang, ``Domain randomization-enhanced depth simulation and restoration for perceiving and grasping specular and transparent objects,'' in \emph{Proc. Eur. Conf. Comput. Vis.(ECCV)}, pp. 374--391, 2022.

\bibitem{tang2021depthgrasp}
Y.~Tang, J.~Chen, Z.~Yang, Z.~Lin, Q.~Li, and W.~Liu, ``Depthgrasp: Depth completion of transparent objects using self-attentive adversarial network with spectral residual for grasping,'' in \emph{IEEE Int. Conf. Intell. Rob. Syst.(IROS)}, pp. 5710--5716, 2021.

\bibitem{li2023fdct}
T.~Li, Z.~Chen, H.~Liu, and C.~Wang, ``Fdct: Fast depth completion for transparent objects,'' \emph{IEEE Rob. Auto. Lett.}, vol.~8, no.~9, pp. 5823--5830, 2023.

\bibitem{zhu2021rgb}
L.~Zhu, A.~Mousavian, Y.~Xiang, H.~Mazhar, J.~van Eenbergen, S.~Debnath, and D.~Fox, ``Rgb-d local implicit function for depth completion of transparent objects,'' in \emph{Proc. IEEE Conf. Comput. Vis. Pattern. Recognit.(CVPR)}, pp. 4649--4658, 2021.

\bibitem{zhai2024tcrnet}
D.-H. Zhai, S.~Yu, W.~Wang, Y.~Guan, and Y.~Xia, ``Tcrnet: Transparent object depth completion with cascade refinements,'' \emph{IEEE Trans. Autom. Sci. Eng.}, vol.~22, pp. 1893--1912, 2025.

\bibitem{yan2024transparent}
Y.~Yan, H.~Tian, K.~Song, Y.~Li, Y.~Man, and L.~Tong, ``Transparent object depth perception network for robotic manipulation based on orientation-aware guidance and texture enhancement,'' \emph{IEEE Trans. Instrum. Meas.}, vol.~73, pp. 1--11, 2024.

\bibitem{jiang2022a4t}
J.~Jiang, G.~Cao, T.-T. Do, and S.~Luo, ``A4t: Hierarchical affordance detection for transparent objects depth reconstruction and manipulation,'' \emph{IEEE Rob. Auto. Lett.}, vol.~7, no.~4, pp. 9826--9833, 2022.

\bibitem{oliva2007role}
A.~Oliva and A.~Torralba, ``The role of context in object recognition,'' \emph{Trends in cognitive sciences}, vol.~11, no.~12, pp. 520--527, 2007.

\bibitem{eigen2014depth}
D.~Eigen, C.~Puhrsch, and R.~Fergus, ``Depth map prediction from a single image using a multi-scale deep network,'' \emph{Adv. Neur. Info. Proc. Syst.(NeurlPS)}, vol.~27, 2014.

\bibitem{fu2018deep}
H.~Fu, M.~Gong, C.~Wang, K.~Batmanghelich, and D.~Tao, ``Deep ordinal regression network for monocular depth estimation,'' in \emph{Proc. IEEE Conf. Comput. Vis. Pattern. Recognit.(CVPR)}, pp. 2002--2011, 2018.

\bibitem{silberman2012indoor}
N.~Silberman, D.~Hoiem, P.~Kohli, and R.~Fergus, ``Indoor segmentation and support inference from rgbd images,'' in \emph{Proc. Eur. Conf. Comput. Vis.(ECCV)}, pp. 746--760, 2012.

\bibitem{geiger2013vision}
A.~Geiger, P.~Lenz, C.~Stiller, and R.~Urtasun, ``Vision meets robotics: The kitti dataset,'' \emph{Int. J. Robot. Res.}, vol.~32, no.~11, pp. 1231--1237, 2013.

\bibitem{han2015fixed}
K.~Han, K.-Y.~K. Wong, and M.~Liu, ``A fixed viewpoint approach for dense reconstruction of transparent objects,'' in \emph{Proc. IEEE Conf. Comput. Vis. Pattern. Recognit.(CVPR)}, pp. 4001--4008, 2015.

\bibitem{qian20163d}
Y.~Qian, M.~Gong, and Y.~H. Yang, ``3d reconstruction of transparent objects with position-normal consistency,'' in \emph{Proc. IEEE Conf. Comput. Vis. Pattern. Recognit.(CVPR)}, pp. 4369--4377, 2016.

\bibitem{xu2021seeing}
H.~Xu, Y.~R. Wang, S.~Eppel, A.~Aspuru-Guzik, F.~Shkurti, and A.~Garg, ``Seeing glass: joint point cloud and depth completion for transparent objects,'' \emph{arXiv preprint arXiv:2110.00087}, 2021.

\bibitem{fan2024tdcnet}
X.~Fan, C.~Ye, A.~Deng, X.~Wu, M.~Pan, and H.~Yang, ``Tdcnet: Transparent objects depth completion with cnn-transformer dual-branch parallel network,'' \emph{arXiv preprint arXiv:2412.14961}, 2024.

\bibitem{liu2021swin}
Z.~Liu, Y.~Lin, Y.~Cao, H.~Hu, Y.~Wei, Z.~Zhang, S.~Lin, and B.~Guo, ``Swin transformer: Hierarchical vision transformer using shifted windows,'' in \emph{Proc. IEEE Int. Conf. Comput. Vis.(ICCV)}, pp. 10\,012--10\,022, 2021.

\bibitem{rombach2022high}
R.~Rombach, A.~Blattmann, D.~Lorenz, P.~Esser, and B.~Ommer, ``High-resolution image synthesis with latent diffusion models,'' in \emph{Proc. IEEE Conf. Comput. Vis. Pattern. Recognit.(CVPR)}, pp. 10\,684--10\,695, 2022.

\bibitem{ranftl2020towards}
R.~Ranftl, K.~Lasinger, D.~Hafner, K.~Schindler, and V.~Koltun, ``Towards robust monocular depth estimation: Mixing datasets for zero-shot cross-dataset transfer,'' \emph{IEEE Trans. Pattern Anal. Mach. Intell.}, vol.~44, no.~3, pp. 1623--1637, 2020.

\bibitem{yin2023metric3d}
W.~Yin, C.~Zhang, H.~Chen, Z.~Cai, G.~Yu, K.~Wang, X.~Chen, and C.~Shen, ``Metric3d: Towards zero-shot metric 3d prediction from a single image,'' in \emph{Proc. IEEE Int. Conf. Comput. Vis.(ICCV)}, pp. 9043--9053, 2023.

\bibitem{yang2024depth}
L.~Yang, B.~Kang, Z.~Huang, Z.~Zhao, X.~Xu, J.~Feng, and H.~Zhao, ``Depth anything v2,'' \emph{Proc. Adv. Neural Inf. Process. Syst.(NeurlPS)}, vol.~37, pp. 21\,875--21\,911, 2024.

\bibitem{lenz2015deep}
I.~Lenz, H.~Lee, and A.~Saxena, ``Deep learning for detecting robotic grasps,'' \emph{Int. J. Robot. Res.}, vol.~34, pp. 705--724, 2015.

\bibitem{wu2021real}
Y.~Wu, F.~Zhang, and Y.~Fu, ``Real-time robotic multi-grasp detection using anchor-free fully convolutional grasp detector,'' \emph{IEEE Trans. Ind. Electron.}, pp. 13\,171--13\,181, 2021.

\bibitem{shang2020deep}
W.~Shang, F.~Song, Z.~Zhao, H.~Gao, S.~Cong, and Z.~Li, ``Deep learning method for grasping novel objects using dexterous hands,'' \emph{IEEE Trans. Cybern.}, vol.~52, pp. 2750--2762, 2020.

\bibitem{dong2024robotic}
H.~Dong, J.~Zhou, and H.~Yu, ``Robotic grasps of cylindrical and cubic objects via real-time learning-based shape detection,'' \emph{IEEE Trans. Autom. Sci. Eng.}, vol.~22, pp. 9681--9697, 2025.

\bibitem{laili2022custom}
Y.~Laili, Z.~Chen, L.~Ren, X.~Wang, and M.~J. Deen, ``Custom grasping: A region-based robotic grasping detection method in industrial cyber-physical systems,'' \emph{IEEE Trans. Autom. Sci. Eng.}, vol.~20, pp. 88--100, 2022.

\bibitem{qi2017pointnet}
C.~R. Qi, H.~Su, K.~Mo, and L.~J. Guibas, ``Pointnet: Deep learning on point sets for 3d classification and segmentation,'' in \emph{Proc. IEEE Conf. Comput. Vis. Pattern. Recognit.(CVPR)}, pp. 652--660, 2017.

\bibitem{qi2017pointnet++}
C.~R. Qi, L.~Yi, H.~Su, and L.~J. Guibas, ``Pointnet++: Deep hierarchical feature learning on point sets in a metric space,'' \emph{Adv. Neur. Info. Proc. Syst.(NeurlPS)}, vol.~30, 2017.

\bibitem{liang2019pointnetgpd}
H.~Liang, X.~Ma, S.~Li, and G{\"o}rner, ``{PointNetGPD}: Detecting grasp configurations from point sets,'' in \emph{IEEE Int. Conf. Rob. Autom.(ICRA)}, pp. 3629--3635, 2019.

\bibitem{fang2020graspnet}
H.-S. Fang, C.~Wang, M.~Gou, and C.~Lu, ``Graspnet-1billion: A large-scale benchmark for general object grasping,'' in \emph{Proc. IEEE Conf. Comput. Vis. Pattern. Recognit.(CVPR)}, pp. 11\,444--11\,453, 2020.

\bibitem{sundermeyer2021contact}
M.~Sundermeyer, A.~Mousavian, R.~Triebel, and D.~Fox, ``Contact-graspnet: Efficient 6-dof grasp generation in cluttered scenes,'' in \emph{IEEE Int. Conf. Rob. Autom.(ICRA)}, pp. 13\,438--13\,444, 2021.

\bibitem{mousavian20196}
A.~Mousavian, C.~Eppner, and D.~Fox, ``6-dof graspnet: Variational grasp generation for object manipulation,'' in \emph{Proc. IEEE Int. Conf. Comput. Vis.(ICCV)}, pp. 2901--2910, 2019.

\bibitem{wang2021graspness}
C.~Wang, H.-S. Fang, M.~Gou, H.~Fang, J.~Gao, and C.~Lu, ``Graspness discovery in clutters for fast and accurate grasp detection,'' in \emph{Proc. IEEE Int. Conf. Comput. Vis.(ICCV)}, pp. 15\,964--15\,973, 2021.

\bibitem{cheng2025pcf}
Y.~Cheng, F.~Zha, W.~Guo, P.~Wang, C.~Zeng, L.~Sun, and C.~Yang, ``Pcf-grasp: Converting point completion to geometry feature to enhance 6-dof grasp,'' \emph{arXiv preprint arXiv:2504.16320}, 2025.

\bibitem{liu2024transparent}
B.~Liu, H.~Li, Z.~Wang, and T.~Xue, ``Transparent depth completion using segmentation features,'' \emph{ACM Trans. Multimedia Comput. Commun. Appl.}, vol.~20, no.~12, pp. 1--19, 2024.

\bibitem{chen2023tode}
K.~Chen, S.~Wang, B.~Xia, D.~Li, Z.~Kan, and B.~Li, ``Tode-trans: Transparent object depth estimation with transformer,'' in \emph{IEEE Int. Conf. Rob. Autom.(ICRA)}, pp. 4880--4886, 2023.

\bibitem{he2016deep}
K.~He, X.~Zhang, S.~Ren, and J.~Sun, ``Deep residual learning for image recognition,'' in \emph{Proc. IEEE Conf. Comput. Vis. Pattern. Recognit.(CVPR)}, pp. 770--778, 2016.

\bibitem{kirillov2023segment}
A.~Kirillov, E.~Mintun, N.~Ravi, H.~Mao, C.~Rolland, L.~Gustafson, T.~Xiao, S.~Whitehead, A.~C. Berg, W.-Y. Lo \emph{et~al.}, ``Segment anything,'' in \emph{Proc. IEEE Int. Conf. Comput. Vis.(ICCV)}, pp. 4015--4026, 2023.

\end{thebibliography}

\end{document}